\documentclass[letterpaper]{article} % DO NOT CHANGE THIS
\usepackage{aaai2026}
\usepackage{times}  % DO NOT CHANGE THIS
\usepackage{helvet}  % DO NOT CHANGE THIS
\usepackage{courier}  % DO NOT CHANGE THIS
\usepackage[hyphens]{url}  % DO NOT CHANGE THIS
\usepackage{graphicx} % DO NOT CHANGE THIS
\usepackage{natbib}  % DO NOT CHANGE THIS AND DO NOT ADD ANY OPTIONS TO IT
\usepackage{caption} % DO NOT CHANGE THIS AND DO NOT ADD ANY OPTIONS TO IT
\usepackage{algorithm}
\usepackage{algorithmic}

\usepackage{newfloat}
\usepackage{listings}

\usepackage{amsmath}
\usepackage{amssymb}

\usepackage{multirow}
\usepackage{booktabs}
\usepackage{array}
\usepackage{adjustbox}
\usepackage{enumitem}

\DeclareCaptionStyle{ruled}{labelfont=normalfont,labelsep=colon,strut=off} % DO NOT CHANGE THIS
\floatstyle{ruled}
\newfloat{listing}{tb}{lst}{}
\floatname{listing}{Listing}
\nocopyright %-- Your paper will not be published if you use this command
\title{Cross-Granularity Hypergraph Retrieval-Augmented Generation \\ for Multi-hop Question Answering}

\author {
    Changjian Wang\textsuperscript{\rm 1,\rm 2},
    Weihong Deng\textsuperscript{\rm 1},
    Weili Guan\textsuperscript{\rm 2},
    Quan Lu\textsuperscript{\rm 1},
    Ning Jiang\textsuperscript{\rm 1}
}
\affiliations {
    \textsuperscript{\rm 1}Mashang Consumer Finance Co., Ltd.\\
    \textsuperscript{\rm 2}Harbin Institute of Technology (Shenzhen)\\
    \{changjian.wang, weihong.deng, quan.lu, ning.jiang\}@msxf.com, guanweili@hit.edu.cn
}

\usepackage{bibentry}
\begin{document}

\maketitle

\begin{abstract}
Multi-hop question answering (MHQA) requires integrating knowledge scattered across multiple passages to derive the correct answer. Traditional retrieval-augmented generation (RAG) methods primarily focus on coarse-grained textual semantic similarity and ignore structural associations among dispersed knowledge, which limits their effectiveness in MHQA tasks. GraphRAG methods address this by leveraging knowledge graphs (KGs) to capture structural associations, but they tend to overly rely on structural information and fine-grained  word- or phrase-level retrieval, resulting in an underutilization of textual semantics. In this paper, we propose a novel RAG approach called HGRAG for MHQA that achieves cross-granularity integration of structural and semantic information via hypergraphs. Structurally, we construct an entity hypergraph where fine-grained entities serve as nodes and coarse-grained passages as hyperedges, and establish knowledge association through shared entities. Semantically, we design a hypergraph retrieval method that integrates fine-grained entity similarity and coarse-grained passage similarity via hypergraph diffusion. Finally, we employ a retrieval enhancement module, which further refines the retrieved results both semantically and structurally, to obtain the most relevant passages as context for answer generation with the LLM. Experimental results on benchmark datasets demonstrate that our approach outperforms state-of-the-art methods in QA performance, and achieves a 6$\times$ speedup in retrieval efficiency.

\end{abstract}


\section{Introduction}

Retrieval-augmented generation (RAG)~\cite{lewis2020retrieval,gao2023retrieval} have emerged as a promising approach to enhance the capabilities of large language models (LLMs).
By integrating external knowledge sources, RAG enables LLMs to access up-to-date and domain-specific information that is not contained within their static parameters. This augmentation significantly improves the accuracy and reliability of the generated responses, addressing the limitations of hallucination~\cite{huang2025survey} and outdated knowledge commonly observed in standalone LLMs.

Despite the achievements of RAG methods, their vector similarity retrieval manner still struggles with knowledge-intensive tasks, such as multi-hop question answering (MHQA), which require knowledge integration across passages. To tackle the drawbacks of traditional RAG, some iterative retrieval methods have been proposed, such as IRCoT~\cite{trivedi2023interleaving} which interleaves retrieval with chain-of-thought (CoT)~\cite{wei2022chain} reasoning to improve the performance on the MHQA task. However, the low efficiency and high response latency of iterative retrieval seriously undermine the user experience. Graph retrieval-augmented generation (GraphRAG) methods~\cite{edge2024local,guo2024lightrag,jimenez2024hipporag},  which establish the association of scattered knowledge through the knowledge graphs (KGs), have received widespread attention in recent years. Instead of directly retrieving text chunks, GraphRAGs retrieve  nodes, edges, or subgraph communities in KGs, and expand information through the graph structure, thereby enhancing the performance of retrieval and generation.

\begin{figure}[t]
    \centering
    \includegraphics[width=0.45\textwidth]{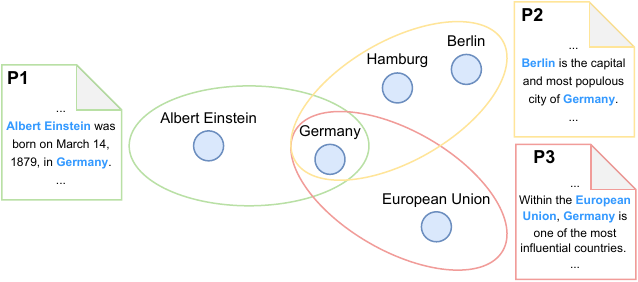}
    \caption{An example of passage association via entities. P1, P2, and P3 are passages about Albert Einstein, Germany, and the European Union, respectively.}
    \label{fig:mt}
\end{figure}

From the perspective of retrieval granularity, traditional RAGs retrieve coarse-grained text chunks, which focus on semantic similarity but ignore the structural association of text; GraphRAGs retrieve fine-grained words or phrases, which rely heavily on the graph structure, leading to insufficient utilization of textual semantics, and the deficiencies are further amplified when the graph structure contains omissions or errors. The fragmented utilization of granularity information in existing methods overlooks the complementarity between different levels of granularity, making it difficult to effectively combine structural and semantic information. However, integrating cross-granularity structural and semantic information can effectively support the MHQA task. As illustrated in Figure~\ref{fig:mt}, we establish direct structural connections between coarse-grained passages via shared fine-grained entities as intermediaries. For a given multi-hop question such as ``What is the capital of the country where Albert Einstein was born?'', leveraging fine-grained entity semantic similarity, we can match the entity ``Albert Einstein'' and quickly locate passage P1. Then we use the shared entity ``Germany'' to find passages P2 and P3 through structural association. By further integrating fine-grained entity similarity and coarse-grained passage similarity, we can determine P1 and P2 (P2 is more similar to the question than P3) as the most relevant passages. We note that a hypergraph, a generalization of graphs whose edges can connect any number of nodes, can naturally model high-order structures for the MHQA task. If we treat entities as nodes and passages as hyperedges, a hypergraph can effectively model structural associations across different granularities through its inherent structure, while also facilitating semantic information propagation via hypergraph diffusion.

In this paper, we propose HGRAG, a hypergraph-based RAG method for MHQA, which enables cross-granularity integration from both structural and semantic perspectives. Structurally, we construct an entity hypergraph by first extracting entities from each passage and then treating entities as nodes and passages containing multiple entities as hyperedges. Leveraging the hypergraph, scattered passages can be connected via shared entities, thereby establishing structural associations between fine-grained entities and coarse-grained passages. Semantically, we propose a hypergraph retrieval method based on hypergraph diffusion. For each query, we construct an entity similarity vector (between query entities and corpus entities) and a passage similarity vector (between query text and passage texts). The passage similarity vector is used to form a passage-weighted hypergraph Laplacian which serves as the diffusion operator applied to the entity similarity vector. Through this process, fine-grained entity-level and coarse-grained passage-level semantic similarities are naturally integrated. Furthermore, we introduce a retrieval enhancement module to refine the hypergraph retrieval results both semantically and structurally, aiming to obtain higher-quality related passages as context for LLM answer generation. We evaluate our proposed method for MHQA on three benchmark datasets, and the experimental results show that our method achieves superior performance in both QA performance and retrieval efficiency.

In summary, our contributions are as follows:
\begin{itemize}[noitemsep,topsep=0pt]
    \item We propose HGRAG, a novel hypergraph-based RAG method for MHQA. HGRAG achieves cross-granularity integration of structure and semantics through entity hypergraph construction, hypergraph retrieval, and retrieval enhancement modules.
    \item We evaluate our method on three benchmark datasets: HotpotQA, 2WikiMultiHopQA, and MuSiQue. Experimental results demonstrate that HGRAG significantly and consistently outperforms state-of-the-art methods.
    \item We compare the retrieval efficiency of HGRAG with the state-of-the-art HippoRAG 2. The results show that our approach not only significantly reduces redundant nodes (by 40\%) and edges, but also achieves a 6× speedup.

\end{itemize}

\section{Related Work}

RAG enhances LLMs by integrating retrieved knowledge and has been widely adopted in various NLP tasks. Traditional RAG methods employ retrievers to obtain texts most relevant to a given query, and use them as context for answer generation by LLMs. Early RAG commonly relied on lightweight dense retrievers~\cite{izacard2022unsupervised,santhanam2022colbertv2}, which embed both queries and passages into a vector space and retrieve based on semantic vector similarity. These dense retrievers have shown better performance than sparse retrievers like BM25~\cite{robertson1994some}. Recent LLM-based retrievers~\cite{li2023towards,muennighoff2025generative,lee2025nvembed} further improve retrieval quality by leveraging richer semantic and contextual representations. However, these methods primarily focus on textual semantics while neglecting structural association, which is crucial for knowledge-intensive tasks like MHQA.

To enhance RAG's ability to associate disparate knowledge, several structure-augmented methods leveraging tree~\cite{sarthi2024raptor} or graph~\cite{peng2024graph} have been proposed. Among them, GraphRAGs based on KGs are the most widely adopted. Graph RAG~\cite{edge2024local} leverages LLMs to construct KGs and generate community summaries over detected multi-layer communities to answer global queries. LightRAG~\cite{guo2024lightrag} simplifies Graph RAG with dual-level retrieval strategies on KGs to construct a fast and scalable RAG system. Inspired by human memory, HippoRAG~\cite{jimenez2024hipporag} employs the Personalized PageRank (PPR)~\cite{haveliwala2002topic} algorithm on KGs to identify passages associated with important entities. Building upon HippoRAG, HippoRAG 2~\cite{gutierrez2025rag} introduces passage nodes and improves the linking strategy to propose a non-parametric continual learning framework. However, these methods heavily rely on graph structures constructed from KGs, while underutilizing coarse-grained textual semantics. Moreover, their KGs consist of triples obtained from open information extraction, which is prone to inaccuracies and omissions. The resulting noisy or incomplete graph structures can further degrade the effectiveness of GraphRAG methods.

Several iterative retrieval methods leverage subquestion decomposition~\cite{press2023measuring}, CoT-based intermediate query generation~\cite{trivedi2023interleaving}, or KGs~\cite{liang2024kag} for MHQA, requiring multiple rounds of LLM inference and retrieval. In addition, some non-RAG methods~\cite{panda2024holmes,li2023leveraging} also leverage KGs or triples to introduce structure information for MHQA. Orthogonal to these approaches, we follow a non-iterative retrieval RAG paradigm, which achieves competitive performance while significantly reducing latency~\cite{jimenez2024hipporag}. We note that some prior studies~\cite{luo2025hypergraphrag,feng2025hyper} introduce hypergraphs into the RAG framework to model n-ary relations for domain-specific tasks. In contrast, our method leverages hypergraphs to model cross-granularity interactions between entities and passages for MHQA. Furthermore, unlike their separate retrieval strategies, our hypergraph diffusion retrieval method integrates cross-granularity semantics in a unified manner.

\begin{figure*}[t]
    \centering
    \includegraphics[width=1.0\textwidth]{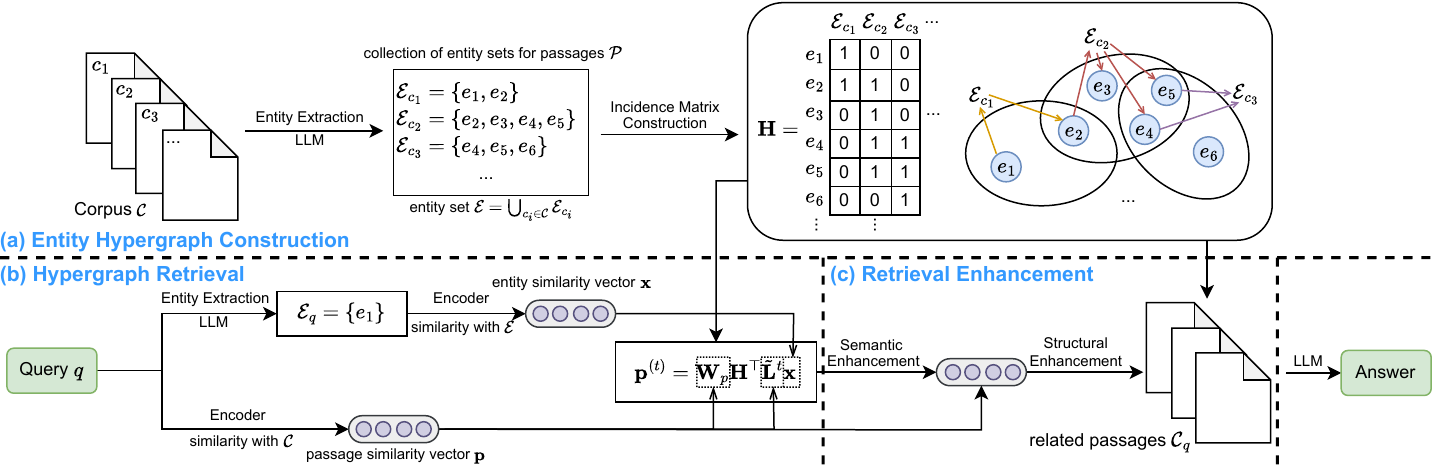}
    \caption{The overview of HGRAG. For a corpus $\mathcal{C}$ composed of multiple passages, in module (a), we extract entities for each passage using an LLM , and construct an entity hypergraph, which views entities as nodes and passages containing multiple entities as hyperedges. Given a query $q$, in module (b), we construct two semantic similarity vectors: $\mathbf{x}$ (entities in $q$ to entities in $\mathcal{E}$) and $\mathbf{p}$ ($q$'s text to passages in $\mathcal{C}$). $\mathbf{x}$ is used as the initial vector for hypergraph diffusion, while $\mathbf{p}$ is used to construct the hypergraph weight matrix and the passage-weighted hypergraph Laplacian. After $t$ steps of diffusion according to Eq.~\ref{eq:hd}, we obtain a new passage relevance vector $\mathbf{p}^{(t)}$, which integrates both entity-level and passage-level semantic similarities. As shown in the illustrated hypergraph, the diffusion process follows an entity--passage--entity pattern, starting from entity $e_1$ that is most similar to the entities in $q$ (arrows in different colors indicating different diffusion steps). In module (c), $\mathbf{p}^{(t)}$ is combined with $\mathbf{p}$ for semantic enhancement. Then, guided by hyperedge associations, structural enhancement is applied to identify the most relevant passages $\mathcal{C}_q$.  Finally, $\mathcal{C}_q$ are used as context for the LLM to generate the answer.}
    \label{fig:arc}
\end{figure*}

\section{Problem Definition}
The multi-hop question answering task is to derive the correct answer to a given question by reasoning over multiple pieces of information from a provided corpus. Formally, given a natural language query $q$ and a corpus $\mathcal{C} = \{ c_i \mid i = 1, 2, \dots, n \}$ which is a set consisting of multiple passages, the task aims to obtain the answer $\mathcal{A}_q$ of $q$ according to $\mathcal{C}$, i.e., $\mathcal{A}_q = f(q, \mathcal{C})$, where $f$ is typically implemented by an LLM in state-of-the-art approaches. In real-world scenarios, the corpus $\mathcal{C}$ is usually large, making it impractical to feed the entire corpus into an LLM due to the input token limits, high computational cost, and the potential introduction of irrelevant information (noise). Therefore, MHQA tasks typically employ a retrieval step to obtain a subset $\mathcal{C}_q = \operatorname{Retrieve}(q, \mathcal{C}) \subseteq \mathcal{C}$ consisting of passages highly relevant to the query $q$. $\mathcal{C}_q$ is used to construct a context for the LLM to generate the final answer:
\begin{equation}
    \mathcal{A}_q = \operatorname{LLM}\left(\operatorname{prompt}(q, \mathcal{C}_q)\right),
\end{equation}
where $\operatorname{prompt}(\cdot)$ is a function that constructs a prompt for the LLM, which typically includes the query, the retrieved passages, and task instructions.

\section{Methodology}
In this section, we will introduce the three components of our method in detail, including the entity hypergraph construction module, the hypergraph retrieval module, and the retrieval enhancement module. Figure~\ref{fig:arc} shows the overview of our method. The entity hypergraph construction module extracts entities from passage texts and constructs a hypergraph with entities as nodes and passages as hyperedges. For a given query, the hypergraph retrieval module employs hypergraph diffusion to integrate entity-level and passage-level similarities, producing refined passage relevance scores. Furthermore, the retrieval enhancement module further improves the retrieval results from the semantic and structural perspective to obtain the most relevant passages for answer generation.

\subsection{Entity Hypergraph Construction}
To associate cross-granularity information from a structural perspective, we construct an entity-centric hypergraph for the corpus, where fine-grained entities are treated as nodes and coarse-grained passages containing multiple entities serve as hyperedges, and passages can be connected via shared entities. Specifically, we first employ an LLM to extract entities from each passage, and then build an entity-passage incidence matrix to represent the hypergraph structure.

\subsubsection{Entity Extraction}
As fine-grained information, entities can establish structural  associations across different passages. We extract entities from a corpus using an instruction-tuned LLM. Specifically, for a given corpus, i.e., a set of passages $\mathcal{C} = \{ c_i \mid i = 1, 2, \dots, n \}$, we extract entities for each passage in the corpus via an LLM with one-shot prompting (the prompts are shown in Appendix), and construct a collection of entity sets for all passages $\mathcal{P} = \{ \mathcal{E}_{c_i} \mid c_i \in \mathcal{C} \}$, where $\mathcal{E}_{c_i}$ denotes the set of entities contained in passage $c_i$. By taking the union of all entities across the passages, we obtain the global entity set $\mathcal{E} = \bigcup_{c_i \in \mathcal{C}} \mathcal{E}_{c_i} $.

\subsubsection{Incidence Matrix Construction}
In a hypergraph, a hyperedge is defined as a non-empty subset of the node set. We treat each entity as a node and each passage containing multiple entities as a hyperedge. A hypergraph can be represented by an incidence matrix. Specifically, we construct an entity-passage incidence matrix $\mathbf{H}\in\{0,1\}^{|\mathcal{E}|\times |\mathcal{P}|}$ with entries defined as:

\begin{equation}
H_{ij}= \mathbb{I}[e_i\in \mathcal{E}_{c_j}] = \left\{
\begin{array}
    {ll}1, & \mathrm{if~}e_i\in \mathcal{E}_{c_j} \\
    0,     & \mathrm{otherwise}
\end{array}\right.,
\end{equation}
where $\mathbb{I}[\cdot]$ denotes the indicator function. This matrix build the relationship between entities and passages. If an entity appears in a passage, the corresponding entry in the matrix is set to 1, otherwise, it is set to 0.

\subsection{Hypergraph Retrieval}
To fuse cross-granularity information on semantics, we propose a hypergraph retrieval method. This method integrates fine-grained entity-level similarity (between query entities and corpus entities) and coarse-grained passage-level similarity (between query text and passage texts). A passage-weighted hypergraph Laplacian is constructed using the passage semantic similarity, which is used to perform hypergraph diffusion starting from an entity similarity vector. After multiple steps of diffusion, we obtain a passage relevance vector that reflects the relevance between the query and each passage, which incorporates semantic information at both the entity and passage levels.

\subsubsection{Semantic Similarity Vector Construction}
We first construct an entity similarity vector and a passage similarity vector for subsequent hypergraph diffusion.

The entity similarity vector reflects the semantic similarity between entities in the query and entities in the corpus. We first extract the entity set $\mathcal{E}_q$ from the query $q$ using an instruction-tuned LLM. For each query entity $e_q \in \mathcal{E}_q$ and corpus entity $e_i \in \mathcal{E}$, we use a dense encoder $E$ fine-tuned for retrieval to obtain their embeddings $E(e_q)$ and $E(e_i)$. The similarity between entities is then computed using the cosine similarity function $\operatorname{Sim}(\cdot)$. Let $v_i =\max_{e_q \in \mathcal{E}_q} \operatorname{Sim}\left(E(e_q), E(e_i)\right)$ and given a threshold $\eta$, we define the entity similarity vector $\mathbf{x} \in \mathbb{R}^{|\mathcal{E}|}$ with elements $x_i =  v_i \cdot \mathbb{I}[v_i > \eta]$.

The passage similarity vector reflects the semantic similarity between the query and each passage in the corpus. Following the traditional RAG framework, we encode both the query text and passage texts using the encoder $E$, and compute their cosine similarities. Each element of the passage similarity vector $\mathbf{p} \in \mathbb{R}^{|\mathcal{P}|}$ is defined as $p_i = \operatorname{Sim}(E(q), E(c_i))$, where $c_i \in \mathcal{C}$ is the $i$-th passage in the corpus.

\subsubsection{Passage-weighted Hypergraph Laplacian}
The hypergraph Laplacian encodes the structure of a hypergraph and serves as a fundamental operator for diffusion processes. In this section, we construct a passage-weighted Laplacian matrix that incorporates passage similarity for the hypergraph diffusion described in the next section. Specifically, we first construct a diagonal hypergraph weight matrix $\mathbf{W}_p \in \mathbb{R}^{|\mathcal{P}|\times|\mathcal{P}|}$ from the  passage similarity vector $\mathbf{p}$, where each diagonal entry is defined as $\mathbf{W}_p(i, i) = \mathbf{p}(i)$.  Then we combine $\mathbf{W}_p$ with the hypergraph structure to construct the following symmetric normalized hypergraph Laplacian~\cite{zhou2006learning}:
\begin{equation}
    \mathbf{L} = \mathbf{I} - \mathbf{D}_{v}^{-\frac{1}{2}}\mathbf{H}\mathbf{W}_p\mathbf{D}_{e}^{-1}\mathbf{H}^{\top}\mathbf{D}_{v}^{-\frac{1}{2}},
\end{equation}
where $\mathbf{I}$ is the identity matrix. $\mathbf{D}_{v} \in \mathbb{R}^{|\mathcal{E}|\times|\mathcal{E}|}$ is a diagonal matrix of node degrees and it is defined as $\mathbf{D}_{v}(i,i) = \sum_{j} \mathbf{H}(i,j)$, which means the number of passages that an entity appears in. $\mathbf{D}_{e} \in \mathbb{R}^{|\mathcal{P}|\times|\mathcal{P}|}$ is a diagonal matrix of hyperedge degrees, with $\mathbf{D}_{e}(j,j) = \sum_{i} \mathbf{H}(i,j)$ representing the number of entities contained in a passage.

\subsubsection{Hypergraph Diffusion}
Hypergraph diffusion refers to the process of propagating information through a hypergraph, allowing the model  to capture complex and high-order interactions by leveraging hyperedges that simultaneously connect multiple nodes. In our scenario, we use hypergraph diffusion to realize the interaction of cross-granularity semantic information between fine-grained entities and coarse-grained passages. This process starts from the initial entity similarity vector and iteratively propagates information on the hypergraph based on the passage-weighted hypergraph Laplacian.

According to the continuous-time heat kernel graph diffusion formulation $\mathbf{x}^{(t)}=e^{-t\mathbf{L}}\mathbf{x}^{(0)}$, we adopt a discrete-time first-order approximation~\cite{chung1997spectral} $\mathbf{x}^{(t+1)}=(\mathbf{I}-\alpha\mathbf{L})\mathbf{x}^{(t)}$, where $t$ denotes the iteration step and $\alpha$ controls the extent of signal propagation in each iteration. For simplicity, we fix $\alpha=1$. Let $\tilde{\mathbf{L}} = (\mathbf{I} - \mathbf{L}) = \mathbf{D}_{v}^{-\frac{1}{2}}\mathbf{H}\mathbf{W}_p\mathbf{D}_{e}^{-1}\mathbf{H}^{\top}\mathbf{D}_{v}^{-\frac{1}{2}}$ and the entity similarity vector $\mathbf{x}$ as $\mathbf{x}^{(0)}$, the result after $t$ steps of diffusion is:
\begin{equation}
    \mathbf{x}^{(t)} = (\mathbf{I} - \mathbf{L})\mathbf{x}^{(t-1)}=\tilde{\mathbf{L}}^{t}\mathbf{x}.
\end{equation}

After $t$ steps of hypergraph diffusion, we perform another entity-to-passage diffusion to obtain a new passage relevance vector $\mathbf{p}^{(t)}\in \mathbb{R}^{|\mathcal{P}|}$ which reflects the relevance between the query and passages after cross-granularity information fusion:
\begin{equation}
    \mathbf{p}^{(t)} = \mathbf{W}_p\mathbf{H}^{\top}\mathbf{x}^{(t)} = \mathbf{W}_p\mathbf{H}^{\top}\tilde{\mathbf{L}}^{t}\mathbf{x}.
\label{eq:hd}
\end{equation}

Reviewing the above iterative form diffusion process, semantic similarity is propagated in a random walk-like manner. The propagation path follows an entity--passage--entity pattern: in each diffusion step, information flows from  entities (nodes) to passages (hyperedges) that contain them, and then to entities (nodes) in these passages. From a numerical perspective, after multiple multiplications between  Laplacian matrix (containing passage similarity) and entity similarity vector, the resulting passage relevance vector assigns higher values to passages that have higher entity-level and passage-level semantic similarity to the query.

\subsection{Retrieval Enhancement}
In this section, we further refine the hypergraph retrieval results with respect to semantics and structure to obtain high-quality related passages for a given query. These passages along with the query and a task instruction are used to construct a prompt for the LLM, and the final answer is obtained from the LLM's response.

\subsubsection{Semantic Enhancement}
The passage similarity vector $\mathbf{p}$ captures the semantic relevance from dense retrieval. We retain these original semantic retrieval results and combine them with our hypergraph retrieval results to form the final passage relevance vector:
\begin{equation}
    \tilde{\mathbf{p}} = (1-\beta) \cdot \mathbf{p}^{(t)} + \beta \cdot \mathbf{p},
\end{equation}
where the hyperparameter $\beta$ is a balancing parameter that controls the contribution of the original semantic retrieval result. This semantic enhancement strategy resembles a residual connection, which improves robustness and mitigates potential semantic degradation caused by incomplete or deficient graph structures.

\subsubsection{Structural Enhancement}
Our retrieval goal is to obtain a set of high-quality related passages $\mathcal{C}_q$ for a query $q$. The common approach is to select the top-$k$ passages based on similarity scores. However, this approach yields a fixed-size $\mathcal{C}_q$ for each query, which lacks flexibility. The performance of question answering is sensitive to the choice of $k$: a small $k$ may fail to cover all relevant passages, while a large
$k$ may introduce noise and increase the number of input tokens.

Instead of a fixed-size $\mathcal{C}_q$, we propose a dynamic-size $\mathcal{C}_q$ selection mechanism utilizing the hypergraph structure. Specifically, we define a range $[k_1, k_2]$ and obtain two top-$k$ passage sets: $\mathcal{C}_q^{k_1} = \text{Top}(\tilde{\mathbf{p}}, k_1)$ and $\mathcal{C}_q^{k_2} = \text{Top}(\tilde{\mathbf{p}}, k_2)$. Then we filter $\mathcal{C}_q^{k_2}$ by retaining only $\mathcal{C}_q^{k_1}$ and $\mathcal{C}_q^{k_1}$'s first-order hyperedge neighbors (i.e., passages that share entities with $\mathcal{C}_q^{k_1}$ in $\mathcal{C}_q^{k_2}$). More formally, for a query $q$, let $\mathbf{s} = \mathbf{H}^\top \mathbf{H} h(\mathcal{C}_q^{k_1})$ where $\left(h(\mathcal{C}_q^{k_1})\right)_i = \mathbb{I}[c_i \in \mathcal{C}_q^{k_1}]$ is a function that converts $\mathcal{C}_q^{k_1}$ into a multi-hot vector and $s_i$ reflects the number of entities shared between $c_i$ and $\mathcal{C}_q^{k_1}$. The related passage set of $q$ is defined as:
\begin{equation}
    \mathcal{C}_q = \left\{ c_i \mid c_i \in \mathcal{C}_q^{k_2} \land s_i > 0 \right\}.
\end{equation}

\section{Experiments}
\subsection{Experimental Settings}
\subsubsection{Datasets}
To evaluate the performance of our proposed method, we perform experiments on three widely-used MHQA datasets: HotpotQA~\cite{yang2018hotpotqa}, 2WikiMultiHopQA~\cite{ho2020constructing}, and MuSiQue~\cite{trivedi2022musique}. For a fair comparison, we use the subsets of the original datasets following HippoRAG 2~\cite{gutierrez2025rag}, which randomly extracts 1,000 questions and collect all candidate passages  (including supporting and distractor passages) forming a corpus for each dataset.

\subsubsection{Baselines}
To provide a comprehensive evaluation, we select three types of baseline methods following ~\cite{gutierrez2025rag}:

\begin{itemize}%[noitemsep,topsep=0pt]
    \item \textbf{Classic Retrievers.} These include traditional dense and sparse retrieval methods that retrieve relevant passages based solely on query-passage similarity. Representative approaches include BM25~\cite{robertson1994some}, Contriever~\cite{izacard2022unsupervised}, and GTR~\cite{ni2022large}.

    \item \textbf{Large Embedding Models.} These methods leverage LLMs for embedding generation, enabling semantically richer and more context-aware retrieval. We choose three high-performing models: Alibaba-NLP/GTE-Qwen2-7B-Instruct~\cite{li2023towards}, GritLM/GritLM-7B~\cite{muennighoff2025generative}, and NVIDIA/NV-Embedv2~\cite{lee2025nvembed}.

    \item \textbf{Structure-Augmented RAG Methods.} These methods enhance traditional RAGs by integrating structural information. We include four state-of-the-art approaches: RAPTOR, Graph RAG~\cite{edge2024local}, HippoRAG~\cite{jimenez2024hipporag}, and HippoRAG 2~\cite{gutierrez2025rag}.

\end{itemize}

\begin{table}[b] 
\centering
\begin{adjustbox}{max width=\linewidth}
\begin{tabular}{
    >{\centering\arraybackslash}m{0.2cm}
    lcccc
}
\toprule
& Method & MuSiQue & 2Wiki & HotpotQA & Avg \\
\midrule
\multirow{3}{*}{\rotatebox[origin=c]{90}{\small\textit{Classic}}} 
    & BM25 & 43.5 & 65.3 & 74.8 & 61.2 \\
    & Contriever & 46.6 & 57.5 & 75.3 & 59.8 \\
    & GTR (T5-base) & 49.1 & 67.9 & 73.9 & 63.6 \\
\midrule
\multirow{3}{*}{\rotatebox[origin=c]{90}{\small\textit{Large}}} 
    & GTE-Qwen2-7B-Instruct & 63.6 & 74.8 & 89.1 & 75.8 \\
    & GritLM-7B & 65.9 & 76.0 & 92.4 & 78.1 \\
    & NV-Embed-v2 (7B) & 69.7 & 76.5 & 94.5 & 80.2 \\
\midrule
\multirow{3}{*}{\rotatebox[origin=c]{90}{\small\textit{Structure}}} 
    & RAPTOR & 57.8 & 66.2 & 86.9 & 70.3 \\
    & HippoRAG & 53.2 & \underline{90.4} & 77.3 & 73.6 \\
    & HippoRAG 2 & \textbf{74.7} & \underline{90.4} & \textbf{96.3} & \underline{87.1} \\

\midrule
\multirow{1}{*}%{\textit{Ours}} 
    & HGRAG  & \underline{74.1} & \textbf{93.0} & \underline{95.5} & \textbf{87.5} \\
\bottomrule
\end{tabular}
\end{adjustbox}
\caption{Retrieval performance on MHQA benchmarks.}
\label{tab:rtr}
\end{table}

\subsubsection{Metrics}
The performance evaluation of MHQA methods is typically conducted on two subtasks: retrieval and QA. We use Recall@5 to evaluate the retrieval task, which calculates the hit rates based on the top-5 retrieval passages and assesses whether any of the top-5 retrieved passages contain the gold evidence. For the QA task, we adopt two metrics: Exact Match (EM) and F1 score. EM reflects strict correctness by requiring an exact string match with the ground-truth answers, while F1 measures the overlap between predicted and ground-truth answers at the token level.

\subsubsection{Implementation Details}
We use NVEmbed-v2~\cite{lee2025nvembed} as the dense encoder $E$, and Llama-3.3-70B-Instruct~\cite{llama3modelcard} with temperature of 0 as our LLM for entity extraction and answer generation. We fix the $k$ range in structural enhancement with $k_1=5$ and $k_2=10$. The three hyperparameters $\eta$,  $\beta$, and $t$ are chosen using 100 examples from training sets of the respective datasets. More implementation details and optimal hyperparameter settings can be found in the Appendix.

\begin{table*}[t]
\centering
\begin{tabular}{llcccccccc}
\toprule
\multirow{2}{*}{Category} & \multirow{2}{*}{Method} & \multicolumn{2}{c}{MuSiQue} & \multicolumn{2}{c}{2Wiki} & \multicolumn{2}{c}{HotpotQA} & \multicolumn{2}{c}{Avg} \\
\cmidrule(lr){3-4} \cmidrule(lr){5-6} \cmidrule(lr){7-8} \cmidrule(lr){9-10}
& & EM & F1 & EM & F1 & EM & F1 & EM & F1 \\
\midrule
\multirow{4}{*}{\textit{Classic}} 
    & None & 17.6 & 26.1 & 36.5 & 42.8 & 37.0 & 47.3 & 30.4 & 38.7 \\
    & BM25~\cite{robertson1994some} & 20.3 & 28.8 & 47.9 & 51.2 & 52.0 & 63.4 & 40.1 & 47.8 \\
    & Contriever~\cite{izacard2022unsupervised} & 24.0 & 31.3 & 38.1 & 41.9 & 51.3 & 62.3 & 37.8 & 45.2 \\
    & GTR~\cite{ni2022large}  & 25.8 & 34.6 & 49.2 & 52.8 & 50.6 & 62.8 & 41.9 & 50.1 \\
\midrule
\multirow{3}{*}{\textit{Large}} 
    & GTE-Qwen2-7B-Instruct~\cite{li2023towards} & 30.6 & 40.9 & 55.1 & 60.0 & 58.6 & 71.0 & 48.1 & 57.3 \\
    & GritLM-7B~\cite{muennighoff2025generative}         & 33.6 & 44.8 & 55.8 & 60.6 & 60.7 & 73.3 & 50.0 & 59.6 \\
    & NV-Embed-v2 (7B)~\cite{lee2025nvembed}  & 34.7 & 45.7 & 57.5 & 61.5 & \underline{62.8} & 75.3 & 51.7 & 60.8 \\
\midrule
\multirow{4}{*}{\textit{Structure}} 
    & RAPTOR~\cite{sarthi2024raptor}    & 20.7 & 28.9 & 47.3 & 52.1 & 56.8 & 69.5 & 41.6 & 50.2 \\
    & Graph RAG~\cite{edge2024local}      & 27.3 & 38.5 & 51.4 & 58.6 & 55.2 & 68.6 & 44.6 & 55.2 \\
    & HippoRAG~\cite{jimenez2024hipporag}       & 26.2 & 35.1 & 65.0 & 71.8 & 52.6 & 63.5 & 47.9 & 56.8 \\
    & HippoRAG 2~\cite{gutierrez2025rag}  & 37.2 & 48.6 & 65.0 & 71.0 & 62.7 & \underline{75.5} & 55.0 & 65.0 \\
\midrule
\multirow{2}{*}{\textit{Ours}}
    & HGRAG (top-5) & \underline{39.4} & \underline{50.7} & \underline{67.7} & \underline{74.9} & \underline{62.8} & \underline{75.5} & \underline{56.6} & \underline{67.0} \\
    & HGRAG & \textbf{42.2} & \textbf{53.8} & \textbf{70.3} & \textbf{78.3} & \textbf{63.9} & \textbf{76.8} & \textbf{58.8} & \textbf{69.6} \\
\bottomrule
\end{tabular}
\caption{QA performance on MHQA benchmarks. None denotes the performance of the LLM without any retrieved passages.}
\label{tab:qar}
\end{table*}

\subsection{Results}
In this section, we present the performance of our method on both the retrieval and QA subtasks. We report the baseline results from~\cite{gutierrez2025rag} and use the same encoder, LLM, and prompt (instruction and demonstrations) for fair comparison.

\subsubsection{Retrieval Results}
The retrieval results on benchmark datasets are presented in Table~\ref{tab:rtr}. It can be observed that large embedding models outperform classic retrievers on Recall@5. This performance gain can be attributed to their foundation on LLMs, which enable them to capture richer semantic representations than small models. 
The structure-augmented RAG methods outperform classic retrievers, highlighting the importance of incorporating structural information into the retrieval process. However, compared to large embedding models, tree-based (e.g., RAPTOR) and graph-based (e.g., HippoRAG) approaches do not exhibit a clear advantage, which is primarily due to their heavy reliance on structural information while underutilizing semantic information.  Especially on more challenging datasets such as MuSiQue requiring longer reasoning hops, structural modeling is more prone to incompleteness and errors, hindering the retrieval performance of structure-augmented RAG methods. HippoRAG 2 enhances semantic utilization by introducing passage nodes and improves the overall retrieval performance. However, due to the lack of explicit cross-granularity modeling, HippoRAG 2 introduces a large amount of redundant entities and relations, leading to lower retrieval efficiency. Our cross-granularity modeling method HGRAG obtains the best Recall@5 scores on 2Wiki, surpassing HippoRAG 2 by a margin of 2.6\%.  On the MuSiQue and HotpotQA datasets, HGRAG performs comparably to HippoRAG 2, and our method achieves higher retrieval efficiency (see Analysis section for details) and yields higher-quality retrieved passages which is reflected in the following QA task performance.

\subsubsection{QA Results}
Table~\ref{tab:qar} shows the  QA performance of various methods on benchmark datasets, using Llama-3.3-70B-Instruct as the QA reader.
The QA performance of different methods follows a similar trend as in the retrieval task: both large embedding models and structure-augmented RAG methods generally outperform classic retrievers. However, their relative performance varies across different datasets and evaluation metrics, as each method primarily focuses on either semantic or structural information. Our proposed HGRAG consistently outperforms all baseline methods, achieving up to a 10.7\% relative improvement in F1 score on the MuSiQue dataset. The experimental results demonstrate the effectiveness of our cross-granularity modeling in both structural and semantic aspects. For a fair comparison, we also include a variant HGRAG (top-5) without structural enhancement, which uses only the top-5 retrieved passages as context. HGRAG (top-5) still consistently outperforms baseline methods. A detailed ablation study is presented in the following sections. Notably, although strong retrieval performance often contributes to improved QA performance, the correlation between the two is not absolute. For example, RAPTOR achieves a higher Recall@5 than HippoRAG on MuSiQue, but its EM and F1 score is lower. This may be because some methods retrieve fewer relevant passages, but these passages are of higher quality and provide more informative context for answering the question. Compared to HippoRAG 2, although our method obtains slightly lower Recall@5 scores on the MuSiQue and HotpotQA datasets, it still outperforms HippoRAG 2 on QA performance. This suggests that the passages retrieved by our method are of higher quality.

\subsection{Analysis}

\subsubsection{Ablation Study}
To further investigate the effectiveness of our method in integrating semantic and structural information, we conduct an ablation study on the HotpotQA dataset. 

For the semantic ablation, we apply two modifications: (1) replacing the hypergraph weight matrix $\mathbf{W}_p$ with an identity matrix (denoted as HGRAG w/o $\mathbf{W}_p$), and (2) removing the semantic enhancement module (denoted as  HGRAG w/o SE). The retrieval results are reported in Table~\ref{tab:ab_se}. As shown in Table~\ref{tab:ab_se}, the semantic-related hypergraph weight matrix $\mathbf{W}_p$ has a significant impact on our method's performance, with its removal leading to a 26\% drop on Recall@5. The semantic enhancement module also plays an important role,  with its removal resulting in up to a 4.5\% decrease on Recall@5. These results highlight the importance of semantic information in our method. 

For the structural ablation, we use NV-Embed-v2, a structure-free semantic retrieval method, as a baseline. Additionally, we investigate the impact of removing the structural enhancement module from our method by using the top-5 and top-10 retrieved passages as context for answer generation, denoted as HGRAG (top-5) and HGRAG (top-10), respectively. The retrieval and QA results are shown in Table~\ref{tab:ab_st}. HGRAG, HGRAG (top-5), and HGRAG (top-10) all outperform NV-Embed-v2 on Recall@5, Recall@10, and F1 score, with up to a 2\% relative improvement in F1. Furthermore, HGRAG outperforms HGRAG (top-5) and HGRAG (top-10), achieving a maximum F1 relative improvement of 1.7\%. Compared to HGRAG (top-5), HGRAG incorporates more structure-related passages, thereby enhancing QA performance. Compared to HGRAG (top-10), HGRAG achieves better QA performance with fewer passages on an average of 8.7 per query, which demonstrates that HGRAG can reduce redundant passages and improve retrieval quality with lower input token costs. The experimental results validate the effectiveness of integrating structural information within our method.

\begin{table}[t]
    \centering
    %\caption{Semantic ablation results on HotpotQA}
    %\label{tab:ab_se}
    \setlength{\tabcolsep}{10pt}
    \begin{tabular}{lccc}
        \toprule
        Method       & Recall@5      & EM            & F1            \\
        \midrule
        HGRAG w/o $\mathbf{W}_p$ & 69.5          & 56.9          & 68.3          \\
        HGRAG w/o SE   & 91.0          & 60.9          & 73.1          \\
        \midrule
        HGRAG          & \textbf{95.5} & \textbf{63.9} & \textbf{76.8} \\
        \bottomrule
    \end{tabular}
    \caption{Semantic ablation results on HotpotQA.}
    \label{tab:ab_se}
\end{table}

\begin{table}[t]
    \centering
    %\caption{Structural ablation results on HotpotQA}
    %\label{tab:ab_st}
    %\setlength{\tabcolsep}{10pt}
    \begin{tabular}{lccc}
        \toprule
        Method        & Recall@5      & Recall@10     & F1            \\
        \midrule
        NV-Embed-v2   & 94.5          & 97.4          & 75.3          \\
        HGRAG (top-5)   & 95.5          & -             & 75.5          \\
        HGRAG (top-10)  & 95.5          & 98.7          & 76.0          \\
        \midrule
        HGRAG (avg-8.7) & \textbf{95.5} & \textbf{98.7} & \textbf{76.8} \\
        \bottomrule
    \end{tabular}
    \caption{Structural ablation results on HotpotQA.}
    \label{tab:ab_st}
\end{table}

\begin{table}[t]
    \centering
    %\caption{Graph scale comparison on MuSiQue}
    %\label{tab:gs}
    \setlength{\tabcolsep}{2pt}
    \begin{tabular}{lccc}
        \toprule
        Method     & No. of nodes & No. of edges & No. hyperedges \\
        \midrule
        HippoRAG 2 & 96,944       & 1,399,367    & -              \\
        HGRAG      & 57,684       & -            & 11,656         \\
        \bottomrule
    \end{tabular}
    \caption{Graph scale comparison on MuSiQue.}
    \label{tab:gs}
\end{table}

\begin{table}[t]
    \centering
    %\caption{Retrieval time comparison on MuSiQue}
    %\label{tab:rt}
    %\setlength{\tabcolsep}{10pt}
    \begin{tabular}{lccc}
        \toprule
                & HippoRAG 2 & HGRAG & HGRAG (GPU) \\
        \midrule
        Time(s) & 86.3       & 13.7  & 2.0         \\
        \bottomrule
    \end{tabular}
    \caption{Retrieval time comparison on MuSiQue.}
    \label{tab:rt}
\end{table}

\subsubsection{Efficiency Analysis}
To demonstrate the retrieval efficiency of our method, we conduct a comparative analysis with the state-of-the-art method HippoRAG 2 on the MuSiQue dataset. The analysis is conducted from two perspectives: graph scale and retrieval time. 

The comparison of graph scale is shown in Table~\ref{tab:gs}. Since two methods adopt different graph structures, it is difficult to directly compare their complexities. However, considering only the nodes that are common to both structures, the number of nodes required by HGRAG is merely 59.5\% of that in HippoRAG 2. Moreover, unlike HippoRAG 2 which relies on a large number of edges (1,399,367), our method requires only a limited number of hyperedges (11,656) to establish connections among nodes.

For a fair comparison of retrieval time, we only consider the execution time of the core retrieval components of both methods, i.e., the PPR module in HippoRAG 2 and the hypergraph diffusion module in HGRAG. Both are executed on the same Intel Xeon Platinum 8558 CPU. 
Table~\ref{tab:rt} reports the comparison of retrieval time for 1,000 queries. Our method is approximately 6.3$\times$ faster than HippoRAG 2. Since our method is implemented using matrix and vector operations, it is well-suited for parallel acceleration on GPUs. In Table~\ref{tab:rt}, we also report the retrieval time of our method on an NVIDIA H200 GPU, which is 43.2$\times$ faster than the CPU-based implementation of HippoRAG 2. Considering the additional latency introduced by LLM inference required for triple filtering in HippoRAG 2, the actual efficiency gap between the two methods would be even larger.

We further analyze the reasons behind the retrieval efficiency of our method. One reason is that the graph scale of HGRAG is smaller. As mentioned earlier, HGRAG avoids a large number of redundant nodes and edges. Another reason is that HGRAG requires fewer iterations than HippoRAG 2. On the MuSiQue dataset, HippoRAG 2 typically performs around 15 iterations per query on the entire graph, whereas HGRAG requires only 4 fixed iterations on the local hypergraph (this number can be further reduced through techniques such as fast matrix exponentiation and caching). Overall, HGRAG achieves higher retrieval efficiency than HippoRAG 2 with lower structural cost and faster retrieval time.

\section{Conclusion}
In this paper, we propose a novel hypergraph-based RAG method called HGRAG for MHQA, which enables cross-granularity integration of structural and semantic information. HGRAG consists of three key modules: (1) the entity hypergraph construction module, which builds a hypergraph over the corpus to establish structural associations between fine-grained entities and coarse-grained passages; (2) the hypergraph retrieval module, which performs hypergraph diffusion to integrate semantic similarity at both the entity and passage levels; and (3) the retrieval enhancement module, which further refines the retrieval results both semantically and structurally to obtain the most relevant passages for answer generation with the LLM. Experimental results on benchmark datasets demonstrate that our approach outperforms state-of-the-art methods in both QA performance and retrieval efficiency.
    
\bibliography{ref}

\end{document}